\documentclass[10pt,twocolumn,letterpaper]{article}

\usepackage{cvpr}
\usepackage{times}
\usepackage{epsfig}
\usepackage{graphicx}
\usepackage{amsmath}
\usepackage{amssymb}

\usepackage{mathtools}
\usepackage{multirow} % added by me
\usepackage[table,xcdraw]{xcolor} % added by me
\usepackage{algorithm}% added by me
\usepackage{algpseudocode}% added by me
\usepackage{subcaption} % added by me
\cvprfinalcopy % *** Uncomment this line for the final submission

\def\cvprPaperID{1960} % *** Enter the CVPR Paper ID here

\ifcvprfinal\fi
\begin{document}
	
\newcommand{\todo}[1]{\textcolor[rgb]{1,0,0}{#1}}
\newcommand{\Unote}[1]{\textcolor[rgb]{0,1,1}{#1}}
\newcommand{\doublecheck}[1]{\textcolor[rgb]{0,0,1}{#1}}
\newcommand{\keypoint}[1]{\vspace{0.0cm}\noindent\textbf{#1}\quad}
\newcommand{\tnote}[1]{\textcolor[rgb]{1,0,0}{(TH: #1)}}

%%%%%%%%% TITLE
\title{Learning Deep Sketch Abstraction}

\author{Umar Riaz Muhammad$^1$ \quad Yongxin Yang$^2$ \quad Yi-Zhe Song$^1$ \quad Tao Xiang$^1$ \quad Timothy M. Hospedales$^{1,2}$\\
	% For a paper whose authors are all at the same institution,
	% omit the following lines up until the closing ``}''.
	% Additional authors and addresses can be added with ``\and'',
	% just like the second author.
	% To save space, use either the email address or home page, not both
    $^1$SketchX, Queen Mary University of London \quad \quad
    $^2$The University of Edinburgh \\
    {\tt\small \{u.muhammad, yizhe.song, t.xiang\}@qmul.ac.uk, \{yongxin.yang, t.hospedales\}@ed.ac.uk}
}

\maketitle
\thispagestyle{empty}

%%%%%%%%% ABSTRACT
\begin{abstract}
	
	Human free-hand sketches have been studied in various contexts including sketch recognition, synthesis and fine-grained sketch-based image retrieval (FG-SBIR). A fundamental challenge for sketch analysis is to deal with  drastically different human drawing styles, particularly in terms of abstraction level. 
	In this work, we propose the first stroke-level sketch abstraction model based on the insight of sketch abstraction as a process of trading off between the recognizability of a sketch and the number of strokes used to draw it. 
	Concretely, we train a model for abstract sketch generation through reinforcement learning of a stroke removal policy that learns to predict which strokes can be safely removed without affecting recognizability. 
	We show that our abstraction model can be used for various sketch analysis tasks including:  (1) modeling stroke saliency and understanding the decision of sketch recognition models, (2)  synthesizing sketches of variable abstraction for a given category, or reference object instance in a photo, and  (3) training a FG-SBIR model with photos only, bypassing the expensive photo-sketch pair collection step.   
	
\end{abstract}

\section{Introduction}
\label{sec:Introduction}

Sketching is an intuitive process which has been used throughout human history as a communication tool. Due to the recent proliferation of touch-screen devices, sketch is becoming more pervasive: sketches can now be drawn at any time and anywhere  on a smartphone using one's finger.  Consequently sketch analysis has attracted increasing attention from the research community. Various sketch related problems have been studied,  including sketch recognition \cite{eitz2012humans, yu2015sketch, yu2017sketch}, sketch based image retrieval  \cite{eitz2011sketch, hu2013performance, yu2016sketch, song2017deep}, forensic sketch analysis \cite{klare2011matching, ouyang2014cross} and sketch synthesis \cite{sangkloy2016scribbler, ha2017neural, li2017free}.

\begin{figure}
	\centering
	\includegraphics[width=\linewidth]{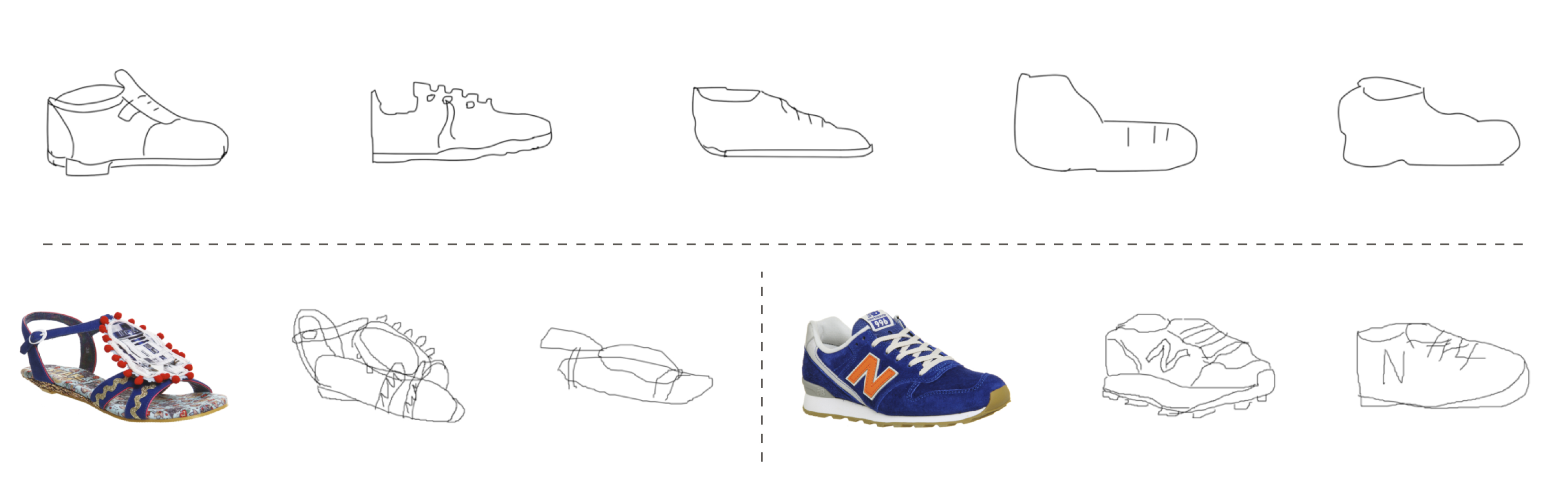}
	\caption{Sketch analysis is difficult because humans draw sketches at very different abstraction levels. Top: different shoe sketches drawn by different people given only the category name. Bottom: sketches are now drawn by different people with a reference photo. }
\vspace{-0.6cm}	\label{fig:SketchAbstractionLevels}
\end{figure}

These studies use free-hand sketches drawn by amateurs based on either a category name, mental recollection, or a reference photo of an object instance.  A fundamental challenge in analyzing free-hand sketches is that sketches drawn by different people for the same object category/instance often differ significantly, especially in their levels of abstraction.   Fig.~\ref{fig:SketchAbstractionLevels} shows some examples of both category-level (drawn with only a category name) and instance-level (drawn with a reference photo) sketches. 
Clearly the large variation in abstraction levels is a challenge for either recognizing the sketch or matching it with a photo. Variation in sketch abstraction level is expected: humans sketch to provide an abstract depiction of an object, and how abstract a sketch is depends both on the task and the individual user's overall and instantaneous preference.
%Such a big variation in sketch abstract level is expected: humans consider sketches as an abstract form of depiction of an  object. However, how abstract a sketch should be is up to the individual drawer's interpretation.  Since there is no way to devise a standard for the definition of abstraction, variable abstraction levels would always be exhibited in sketches. 

We present the first model of deep sketch abstraction.
Our approach to model abstraction is based on the insight that abstraction is a process of tradeoff between recognizability and brevity/compactness (number of strokes). It is thus intuitive that abstraction should vary with task (e.g., sketching for instance- rather than category-level tasks permits less abstraction as the recognition task is more fine-grained), and that abstraction varies between people as their subjective perception (what  seems to be recognizable), as might their relative preference for brevity vs identifiability. 
% The basic assumption is that, even though different humans will have different understandings on how abstract a sketch should be, there is a shared underlying constraint, i.e., the object depicted by a sketch must be recognizable. 
Based on the same insight, we develop a computational model that learns to  abstract concrete input sketches and estimate stroke saliency by finding the most compact subset of input strokes for which the sketch is still recognizable. We consider this similar to the human sketching process: before drawing an object a human has a more detailed mental model of the object, then they work out which details can be safely removed in conveying a compact yet recognizable sketch depiction of the imagined object.

%Based on this assumption,  we develop a computational model to measure the importance of each constituent part of a sketch at two levels of granularities, i.e. strokes that are further decomposed into segments,  in terms of its contribution for making the final sketch recognizable. With such a model, we can progressively remove the least contributing stroke/segment to obtain a more and more abstract sketch whilst minimising the loss in its recognizability. We consider that this process may be similar in spirit to the human sketching process: before drawing a sketch of an object, a human would have gone through a mental abstraction process -- starting with removing colour and texture information, followed by gradually removing more and more contours till it reaches the desirable abstraction level according to the drawer's latent personal standard. 

Specifically, we develop a recurrent neural network (RNN) based abstraction model, which learns to measure the importance of each segment and make a decision on whether to skip or keep it. The impact of any given part removal on recognizability is interdependent with which other parts are kept/removed. We model this dependency as a sequential decision making process. % as per the sequential sketching process. 
Our RNN uses bi-directional gated recurrent units (B-GRU) along with a moving window MLP to capture and extract the contextual information of each sketch-part at each time step. %Because the part-skipping process is sequential ()
%Note that the part-skipping action is sequential and skipping/keeping the current part has effects on the subsequent actions for other parts, which can only be known once the abstraction process completes.
Such a model cannot be learned with conventional supervised learning. We propose a framework for training a sketch abstraction model with reinforcement learning (RL)  using a novel reward scheme that uses the classification rank of the  sketch at each time step to make rewards more informative.
% In this work, a sketch abstraction RL framework is proposed with  a new learning reward scheme. It utilises the classification rank information of the output sketch at each time step to embed saliency information in pre-defined reward values. %At the end after the whole sketch has been processed the survived sketch-parts are fed into a classifier and based on the result of classification a huge positive or negative reward is assigned, which enforces the final sketch to be recognizable. Recurrent Neural Network (RNN), being an obvious choice to deal with sequential data in the form of sketch data-segments, is used to model the agent. The peculiarity of this RNN based agent consists of the use of bi-directional gated recurrent units (B-GRU) along with a multi-layered neural network (moving window) which captures and extracts the contextual information of each sketch-part at each time step.

Using our abstraction model,  we can address a number of problems: 
(1) \textbf{Modeling sketch stroke saliency:} We can estimate stroke saliency as a byproduct of learning to produce brief recognizable sketches. % \doublecheck{thanks to the employed reward scheme which dynamically measures the part saliency information in an abstraction process.}
(2) \textbf{Category-level sketch synthesis with controllable abstraction:}  Given an existing category-level sketch synthesizer, our model can be used to control the level of abstraction in the synthesized sketches.  
(3) \textbf{Instance-level photo-to-sketch synthesis:}  We propose a new approach to photo $\to$ sketch synthesis motivated by human sketching rather than image translation  \cite{sangkloy2016scribbler, isola2016image}. %, a new approach is proposed, motivated by the human sketching process. 
Given a photo, we extract  an edge-map %with appropriate random distortion, 
and treat it as a sketch at the most concrete level. Our sketch abstraction model is then applied to abstract the edge-map into a free-hand style sketch.  
(4) \textbf{FG-SBIR without photo-sketch pairs:}  The photo-to-sketch synthesis model above is used to synthesize photo-freehand sketch pairs using photo input only. This allows us to train an instance-level fine-grained SBIR (FG-SBIR) model without manual data annotation, and moreover it generates data at diverse abstraction levels so the SBIR model is robust to variable abstraction at runtime.% helps train that model to be robust to robust to variable abstraction in human drawing at runtime.
% training a instance-level fine-grained SBIR (FG-SBIR) model. By doing this, the key challenge of collecting sufficient photo-sketch pairs is addressed. Importantly, a FG-SBIR model trained with the synthesised data would be able to cope with variable human drawing styles in term of abstraction level, another major difficulty for  FG-SBIR.  

Our contributions are as follows: 
(1) For the first time, the problem of stroke-level sketch abstraction is studied.
%(2)  A sketch abstraction model  is proposed which is learned in a reinforcement learning framework with a novel reward scheme. 
(2) We propose a reinforcement learning framework with novel reward for training a sketch abstraction model
(3) Both category- and instance-level sketch synthesis can be performed with controllable abstraction level. {We demonstrate that the proposed photo-to-sketch approach is superior than the state-of-the-art alternatives. }
(4) FG-SBIR can now be tackled without the need to collect photo-sketch pairs. Our experiments on two benchmark datasets show that the resulting FG-SBIR model is quite competitive, thus providing the potential to scale FG-SBIR to an arbitrary number of object categories as long as sufficient photos can be collected.  

%-------------------------------------------------------------------------
\section{Related Work}
\label{sec:RelatedWork}

\keypoint{Sketch recognition}
Early work on sketch recognition focused on CAD or artistic drawings \cite{jabal2009comparative, lu2005new, sousa2009geometric}. Inspired by the release of the first large-scale free-hand sketch dataset \cite{eitz2012humans}, subsequent work studied free-hand sketch recognition \cite{eitz2012humans, schneider2014sketch, li2015free} using various hand-crafted features together with classifiers such as SVM.   Yu {\em et al}. \cite{yu2015sketch} proposed the first deep convolutional neural network (CNN) designed for sketch recognition which outperformed previous hand-crafted features by a large margin. In this work we do not directly address sketch recognition. Instead we exploit a sketch recognizer to quantify sketch recognizability and generate recognizability-based rewards to train our abstraction model using RL. In particular, we move away from the conventional CNN modeling of sketches \cite{yu2015sketch, yu2017sketch} where sketches are essentially treated the same as static photos, and employ a RNN-based classifier that fully encodes stroke-level ordering information.

\keypoint{Category-level sketch synthesis} Recently there has been a surge of interest in deep image synthesis   \cite{goodfellow2016nips, sonderby2016ladder, kingma2016improving, reed2017parallel}. Following this trend the first free-hand sketch synthesis model was proposed in \cite{ha2017neural}, which exploits a sequence-to-sequence Variational Autoencoder (VAE). In this model the encoder is a bi-directional RNN that inputs a sketch and outputs a latent vector, and the decoder is an autoregressive RNN that samples output sketches conditioned on a latent vector. %\doublecheck{An unconditional version of the model is also provided consisting of only the decoder module.} 
They combine RNN with Mixture Density Networks (MDN) \cite{graves2013generating} in order to generate continuous data points in a sequential way. In this paper, we use the unconditional synthesizer in \cite{ha2017neural} in conjunction with our proposed abstraction model to synthesize sketches of controllable abstraction level.

\keypoint{Instance-level sketch synthesis}  A sketch can also be synthesized with a reference photo, giving rise to the instance-level sketch synthesis problem. This is an instance of the well studied cross-domain image synthesis problem.  Existing approaches typically adopt  a cross-domain deep encoder-decoder model. 
%Such a model takes an image as input and produces a code/latent feature  embedding using an encoder. A decoder then takes the code as input and generates an image that is related to the input but not identical and is from a different domain. 
Cross-domain image synthesis approaches fall into two broad categories depending on whether the input and output images have pixel-level correspondence/alignment. The first category includes models for super-resolution~\cite{SISR17},  restoration and inpainting \cite{pathak2016context}, which assume pixel-to-pixel alignment. The second category relaxes this assumption and includes models for style transfer (e.g., photo to painting) \cite{PLST16} and cross-domain image-conditioned image generation \cite{PixelTransfer16}.  Photo-to-sketch is extremely challenging due to the large domain gap and the fact that the sketch domain is generated by humans with variable drawing styles. As a result, only sketch-to-photo synthesis has been studied so far \cite{sangkloy2016scribbler, isola2016image, li2017free}. In this work, we study photo-to-sketch synthesis with the novel approach of treating sketch generation as a photo-to-sketch abstraction process. We show that our method generates  more visually appealing sketches than the existing deep cross-domain image translation based approaches such as \cite{sangkloy2016scribbler}.

\keypoint{Sketch based image retrieval} Early effort focused on the category-level SBIR problem \cite{eitz2010evaluation, eitz2011sketch, hu2010gradient, cao2011edgel, cao2010mindfinder, 	wang2010mindfinder, hu2011bag, lin20133d, hu2013performance} whereby a sketch and a photo are considered to be a match as long as they belong to the same category.  %were based on hand-crafted features extraction for both sketches and edge-maps of photos which were then fed into Bag-of-Words (BoW) method to bridge the domain gap. Recently convolutional neural networks (CNNs) have been used \cite{sangkloy2016sketchy, qi2016sketch} to solve this problem achieving much better performance than previous methods.
In contrast, in instance-level fine-grained SBIR (FG-SBIR), they are a match only if they depict the same object instance. FG-SBIR has more practical use, e.g., with FG-SBIR one could use sketch to search to buy a particular shoe s/he just saw on the street \cite{yu2016sketch}.  It has thus received increasing attention recently. State-of-the-art FG-SBIR models  \cite{yu2016sketch, sangkloy2016sketchy}  adopt a multi-branch CNN to learn a joint embedding where photo and sketch domains can be compared. They face two major problems: collecting sufficient matching photo-sketch pairs is tedious and expensive, which severely limits their scalability. In addition, the large variation in abstraction level exhibited in sketches for the same photo (see Fig.~\ref{fig:SketchAbstractionLevels}) also makes the cross-domain matching difficult. In this work, both problems are addressed using the proposed sketch abstraction and  photo-to-sketch synthesis models. 

%.  The problem of FG-SBIR was first introduced in \cite{li2014fine} where this problem was tackled by employing a deformable part-based model (DPM) and graph matching. Recently, solutions based on deep learning approach has been proposed in \cite{yu2016sketch, sangkloy2016sketchy} with the aim of learning a higher-level feature representation and cross-domain matching function. The main difference between \cite{yu2016sketch} and \cite{sangkloy2016sketchy} is that the first uses a Siamese network by taking an extracted edge-map form the photo branch, while the latter uses a heterogeneous network without edge-map extraction. In both works they experimented with pairwise verification and triplet ranking losses, and concluded that the latter performs better. Following this conclusion in our work we use a three-branch CNN model proposed in \cite{yu2016sketch} with triplet ranking loss.

\keypoint{Visual abstraction} The only work on sketch abstraction is that of \cite{berger2013style} where a data-driven approach is used to study style and abstraction in human face sketches. An edge-map is computed and edges are then replaced by similar strokes from a collection of artist sketches.  In contrast, we take a model-based approach and model sketch abstraction from a very different perspective: abstraction is modeled as the process of trading off between compactness and recognizability by progressively removing the least important parts. 
Beyond sketch analysis, visual  abstraction has been studied in the photo domain including  salient region detection \cite{cheng2013efficient}, feature enhancement \cite{kang2009flow}, and low resolution image generation \cite{gerstner2013pixelated}. None of these approaches can be applied to our sketch abstraction problem.

\section{Methodology}
\label{sec:Methodology}

\subsection{Sketch abstraction}
\label{subsec:SketchAbstraction}

\begin{figure*}
	\centering
	\begin{subfigure}[b]{0.49\textwidth}
		\includegraphics[width=\textwidth]{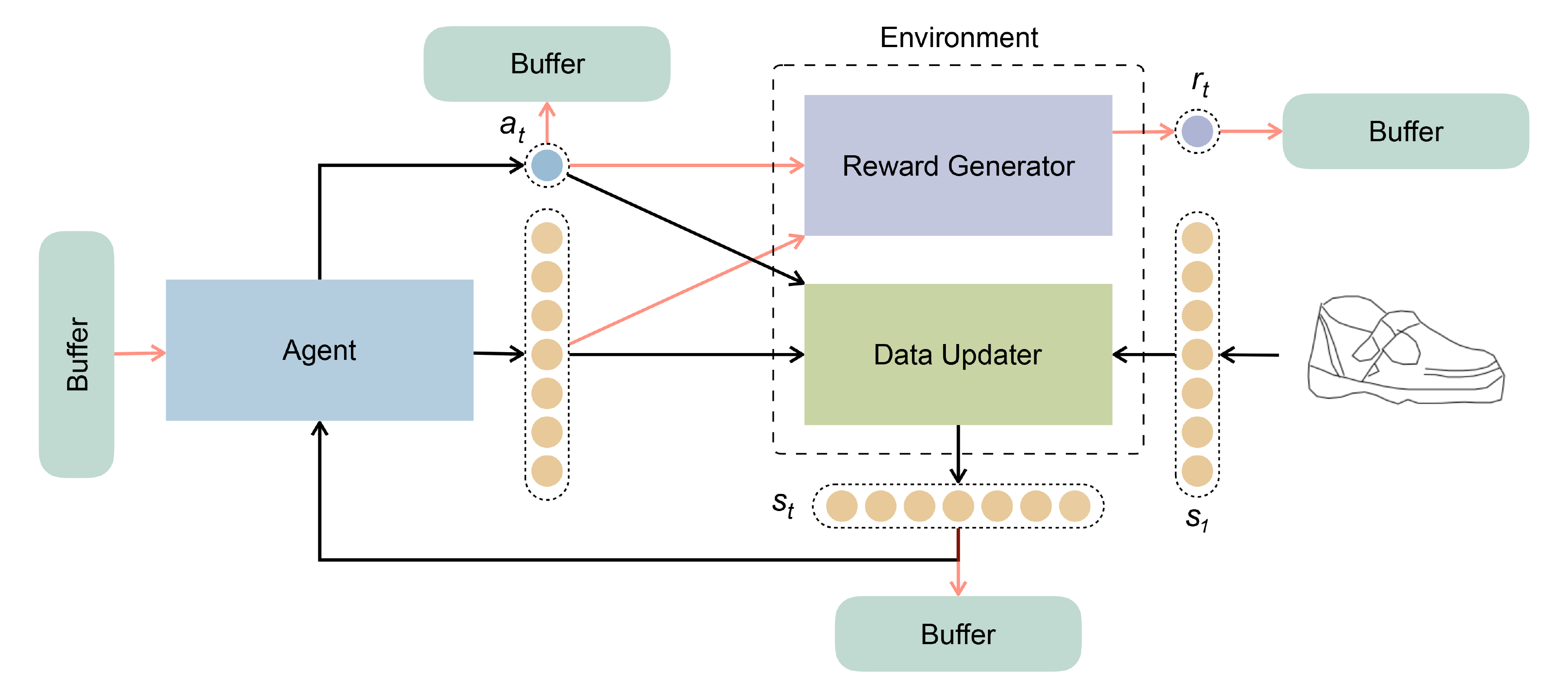}
		\caption{Reinforcement learning framework.}
		\label{fig:RL}
	\end{subfigure}
	\hfill
	\begin{subfigure}[b]{0.49\textwidth}
		\includegraphics[width=\textwidth]{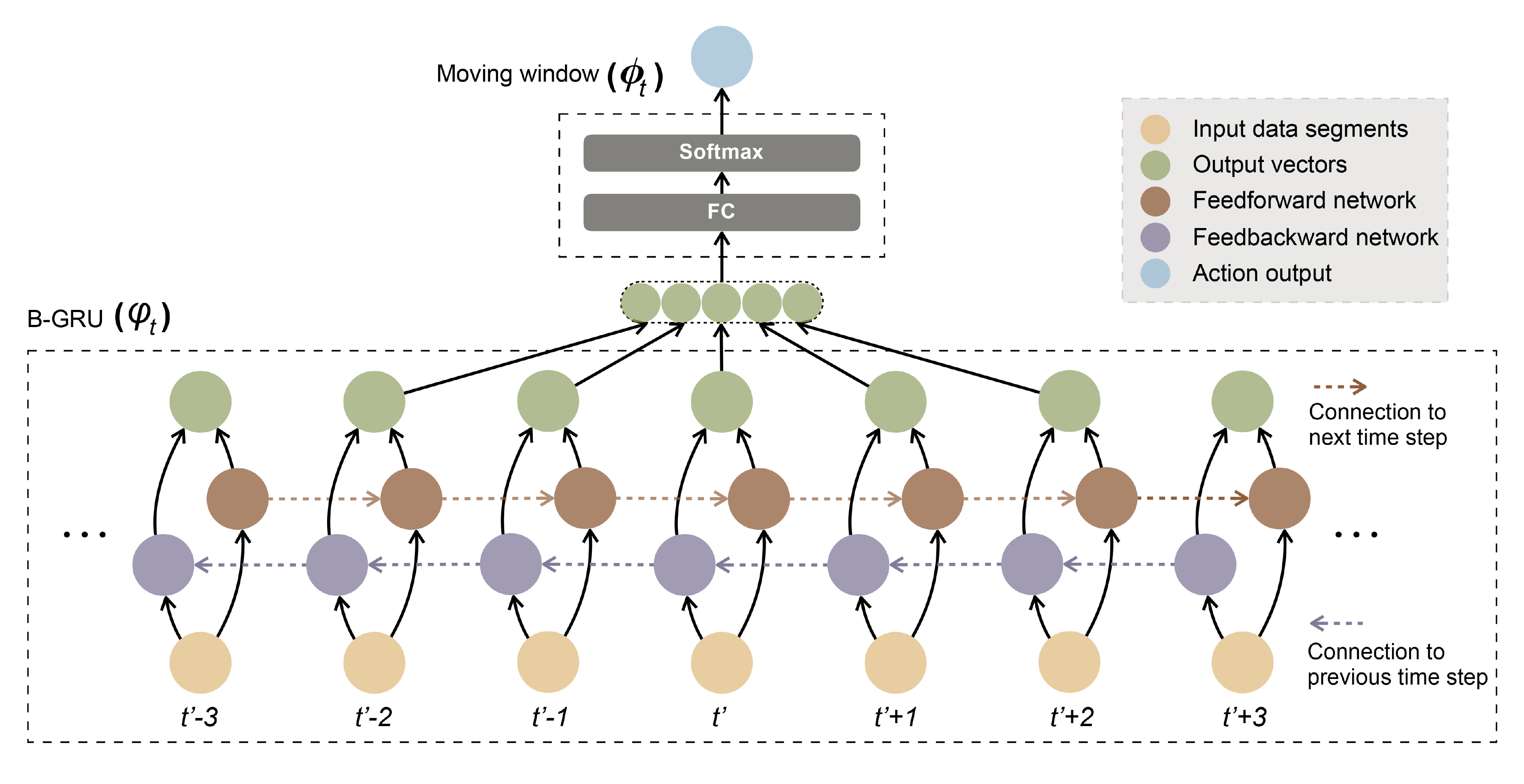}
		\caption{Agent architecture.}
		\label{fig:Agent}
	\end{subfigure}
	\caption{Schematic of our sketch abstraction model.}
	\label{fig:SketchAbstractionModel}
\end{figure*}

\subsubsection{Sketch representation}
\label{subsubsection:SketchRepresenation}
Sketches are represented in a vectorized format. Strokes are encoded as a sequence of coordinates, consisting of 3 elements $(\Delta x, \Delta y, p) $, as in \cite{graves2013generating} for representing human handwriting. We define data-segment as one coordinate and stroke-segment as a group of five consecutive coordinates. Each stroke thus comprises a variable number of stroke-segments.

% This is very important and shouldn't be placed as footnote, because data-segment and stroke-segment are used a lot 

\subsubsection{Problem formulation}
\label{subsubsection:ProblemFormulation}

We formulate the sketch abstraction process as the sequence of decisions made by an abstraction agent which observes stroke-segments in sequence and decides which to keep or remove. The sequence of strokes may come from a model \cite{ha2017neural} when \emph{generating} abstract sketches, or a buffer when \emph{simplifying} an existing human sketch or edge-map. The agent is trained with  reinforcement learning, and  learns to estimate the saliency of each stroke in order to achieve its goal of compactly encoding a recognizable sketch.

The RL framework is described by a Markov Decision Process (MDP), which is  a tuple $\langle \mathcal{S}, \mathcal{A} , \mathcal{T}, \mathcal{R} \rangle$. Here: $\mathcal{S}$ is the set of all possible states, which are observed by the agent in the form of data-segments representing the sketch and the index  pointing at the current stroke-segment being processed.  $\mathcal{A} = \{0, 1\}$ is the set of binary action space representing skipping ($0$) or keeping ($1$) the current stroke-segment. $\mathcal{T}(s_{t+1} | s_{t}, a_{t})$ is the transition probability density from current state $s_{t} \in \mathcal{S}$ to next state $s_{t+1} \in \mathcal{S}$ when the agent takes an action $a_{t} \in \mathcal{A}$. It updates the index and the abstracted sketch so far.  $\mathcal{R}(s_{t}, a_{t}, s_{t+1})$ is the function describing the reward in transitioning from $s_{t}$ to $s_{t+1}$ with action $a_{t}$. 
At each time step $t$, the agent's decision procedure is characterized by a stochastic policy $\pi_{\theta} = \pi(a_{t}|s_{t}, \theta)$ parametrized by $\theta$, which represents the conditional probability of taking action $a_{t}$  in state $s_{t}$.

At first time step $t_{1}$, $s_{1}$ corresponds to the data-segments of the complete sketch with index pointing at the first stroke-segment. The agent evaluates $s_{1}$ and takes an action $a_{1}$ according to its policy $\pi_\theta$, making a decision on whether to keep or skip the first stroke-segment. The transition $\mathcal{T}(s_{2} | s_{1}, a_{1})$ says:  if $a_{1} = 0$ (skip), the next state $s_{2}$ corresponds to the updated data-segments which do not contain the skipped stroke-segment and with the index pointing to next stroke-segment. If $a_{1} = 1$ (keep), the next state $s_{2}$ corresponds to the same data-segments as in $s_{1}$ but with the index pointing to the next stroke-segment. This goes on until the last stroke-segment is reached.

Let $\mathcal{D} = (s_1, a_1,..., s_M,a_M, s_{M+1})$ be a trajectory of length $M$, corresponding to the number of stroke-segments in a sketch. Then the goal of RL is to find the optimal policy $\theta^*$ that maximizes the expected return (cumulative reward discounted by $\gamma \in [0,1]$):

\begin{equation}
	J(\theta) =   \mathbb{E} \left( \sum_{t=1}^{M}  \gamma^{t-1} \mathcal{R}(s_{t}, a_{t}, s_{t+1}) \: \big| \: \pi_{\theta} \right)
	\label{Eq:ExpectedReturn}
\end{equation}
%\noindent where $\gamma \in [0,1]$ is a discount factor.

\subsubsection{Model}
\label{subsubsection:Model}

Our RL-based sketch abstraction model is illustrated in Fig.~\ref{fig:SketchAbstractionModel}(a). A  description of each component follows.

\keypoint{Agent} It consists of  two modules. In the first {\em B-GRU module}, data-segments corresponding to state $s_{t}$ are input sequentially to a recurrent neural network (RNN), i.e., one segment at each time step $t'$ (as shown in Fig.~\ref{fig:SketchAbstractionModel}(b)). We use bi-directional gated recurrent units \cite{DBLP:journals/corr/ChungGCB14} (B-GRU) in the RNN
%. The use of GRU is motivated by its efficiency to process long term patterns in a sequential data just as LSTM (long short term memory) but with a reduced number of parameters, which enables the agent to learn the complex and long sequence relationships more efficiently. While bidirectional network, which consists of neural connections in both forward and backward direction, is used 
to learn and embed past and future information at each time step $t'$. This module represents input data in a compact vectorized format $\varphi_t$ by concatenating the outputs of all time steps. The second {\em moving window module} consists of a multi-layer perceptron (MLP) with two fully-connected layers. The second layer is softmax activated, and generates probabilities for agent actions $\phi_t$. This module slides over the B-GRU module and takes as input those outputs centered at the current stroke-segment under processing, using the index  in  state $s_{t}$. The  architecture of our agent is shown in Fig~\ref{fig:SketchAbstractionModel}(b). 

\keypoint{Environment} The environment implements state transition and reward generation. The state transition module reads the action $a_t$ and state $s_t$ at each time step $t$, and transits the environment to state $s_{t+1}$ by updating data-segments and index of the stroke-segment under processing. In case of a skip action, this update consists of eliminating the skipped data-segments, modifying the rest appropriately given the created gap, and moving the index to the next stroke-segment. In case of a keep action, only the index information is updated. 
The second module is a reward generator which assigns a reward to each state transition. We next describe in detail the proposed reward schemes.

\subsubsection{Reward scheme}
\label{subsubsection:RewardScheme}

We want our agent to abstract sketches by dropping the least important stroke-segments while keeping the final remaining sketch recognizable. Therefore our reward is driven by a sketch recognizability signal deduced from the classification result of a multi-class sketch classifier. 
In accordance with the vectorized sketch format that we use for RL processing, we use a three-layer LSTM \cite{Hochreiter:1997:LSM:1246443.1246450} classifier trained with cross-entropy loss and Adam optimizer \cite{DBLP:journals/corr/KingmaB14}. Using this classifier, we design two types of reward schemes:

\keypoint{Basic reward scheme} This reward scheme is designed to encourage high recognition accuracy of the final abstracted sketch while keeping the minimum number of stroke-segments. For a trajectory of length $M$, the basic reward $b_{t}$ at each time step $t$ is defined as: 
\begin{equation}
	R_{t} =
	b_t = 
	\begin{cases}
		+1,				& \text{if}\ t < M \text{ and}\ a_t = 0~~\text{(skip)}\\
		-5,              & \text{if}\ t < M \text{ and}\ a_t = 1~~ \text{(keep)}\\
		+100		  & \text{if}\ t = M \text{ and Class}(s_t)\ =\ \text{G}\\
		-100		   & \text{if}\ t = M \text{ and Class}(s_t)\ \neq\ \text{G}\\
	\end{cases}
	\label{Eq:ra}
\end{equation}

\noindent where G denotes the ground truth class of the sketch, and Class($s_t$) denotes the prediction of the sketch classifier on abstracted sketch in $s_t$. From Eq.~\ref{Eq:ra}, it is clear that $R_{t}$ is defined to encourage compact/abstract sketch generation (positive reward for skip  and negative reward for keep action), while forcing the final sketch to be still recognizable (large reward if recognized correctly, large penalty if not).

\keypoint{Ranked reward scheme} In this scheme we extend the basic reward by proposing a more elaborate reward computation, aiming to learn the underlying saliency of stroke-segments by integrating the classification rank information at each time step $t$. The total reward is now defined as: 

\begin{equation}
	R_{t} = 
	w_{b} \: b_{t} + w_{r} \: r_{t}
	\label{Eq:r}
\end{equation}
\begin{equation}
	r_{t} = 
	\begin{cases}
		(w_{c} \: c_{t} + w_{v} \: v_{t}) \: b_{t}	& \text{if}\ t < M\\
		0		  										          & \text{if}\ t = M\\
	\end{cases}
	\label{Eq:rRank}
\end{equation}
\begin{small}
	\begin{equation}
		c_{t} = 1 - \left( \frac{\mathrm{K} - \mathit{C_{t}}}{\mathrm{K}} \right) 
	\end{equation}
\end{small}
\begin{small}
	\begin{equation}
		v_{t} = 1 - \left(  \frac{\mathrm{K} - \mathit{(C_{t} - C_{t-1})}}{2 \cdot \mathrm{K}} \right) 
	\end{equation}
\end{small}

\noindent where $r_{t}$ is the ranked reward, $w_{b}$ and $w_{r}$ are weights for the basic  and  ranked reward respectively, $C_t$ is the predicted rank of ground-truth class and $K$ is the number of sketch classes. The \emph{current ranked reward} $c_t$ prefers the ground-truth class to be highly ranked. Thus improving the rank of the ground truth is rewarded even if the classification is not yet correct -- a form of reward-shaping \cite{Wiewiora2010}. The \emph{varied ranked reward} $v_{t}$ is given when the ground-truth class rank improves over time steps. $w_{c}$ and $w_{v}$ are weights for current ranked reward and varied ranked reward respectively. For example, assuming $w_{b} = w_{r} = 0.5$, at time step t, if $a_t = 0$ (skip), then $R_t$ would be $0.5$ when $c_t = 0$, $v_t = 0$, and $R_t=1.0$ when $c_t = 1$, $v_t = 1$; on the other hand if $a_t = 1$ (keep), then $R_t$ would be $-2.5$ when $c_t = 0$, $v_t = 0$, and $R_t=-5.0$ when $c_t = 1$, $v_t = 1$.

The basic vs ranked reward weights $w_{b} \in [0,1]$ and $w_{r} \in [0,1]$ ($w_{b} + w_{r} = 1$) are computed dynamically as a functions of time step $t$. At the first time step $t=1$, $w_{r}$ is 0; subsequently it increases linearly to the fixed final $w_{r_f}$ value at the last time step $t=M$. Weights $w_{c}$ and $w_{v}$ are static with fixed values, such that $w_{c} + w_{v} = 1$.

\vspace{-0.1cm}
\subsubsection{Training procedure}
\label{subsubsection:TrainingProcedure}

We use a policy gradient method to find the optimal policy $\theta^*$ that maximizes the expected return value defined in Eq.~\ref{Eq:ExpectedReturn}. Thus the training consists of sampling the stochastic policy and adjusting the parameters $\theta$ in the direction of  greater expected return via gradient ascent: %TH: It's reward not loss. Ascend it?
\begin{equation}
	\theta \longleftarrow \theta - \eta \: \nabla_\theta J(\theta),
	\label{Eq:GradientDescent}
\end{equation}
\noindent where $\eta$ is the learning rate. In order to have a more robust training, we process multiple trajectories accumulating $\langle s_{t}, a_{t}, R_{t}, s_{t+1} \rangle$ in a Buffer B (see Fig.~\ref{fig:SketchAbstractionModel}(a), and update parameters $\theta$ of the agent every $N$ trajectories. 

\vspace{-0.1cm}
\subsubsection{Controlling abstraction level}
\label{subsubsec:VariedLevelOfAbstraction}

Our trained agent can be used to perform abstraction in a given sketch by  sampling actions $a_t \in \{1, 0\}$ from the agent's output distribution $\phi_t$ in order to keep or skip stroke-segments. We attempt to control the abstraction level by varying the temperature parameter of the softmax function in the  \textit{moving window module} of our agent. However empirically we found out that it does not give the satisfactory result, so instead we  introduce a   shift $\delta$ in the $\phi_t$ distribution to obtain different variants of $\phi_t$, denoted as $\phi^*_t$:
\begin{equation}
	\phi^*_t = \left(  \phi_{t}(a_t=0) + \delta , \: \phi_{t}(a_t=1) - \delta \right)
	\label{Eq:control abstract}
\end{equation}
\noindent where, $\phi_{t}(a_t=0) + \phi_{t}(a_t=1) = 1$ and $\delta \in [-1,1]$. By varying the $\delta$ value we can  obtain arbitrary level of abstraction in the output sketch by biasing towards skip or keep. %Note that although we can  abstract a sketch by removing its stroke-segments, in practice, to make the remaining strokes smooth and avoid fragmentation, the skip/keep action is applied to each stroke instead. This is done in a post-processing step that aggregates the actions of the constituent stroke-segments of each stroke by majority voting. \todo{double check} %Comment by Umar: Actually we do this only for egde-map abstraction images used for FG-SBIR% 
The code for our abstraction model will be made available from the SketchX website: \url{http://sketchx.eecs.qmul.ac.uk/downloads/}.

\subsection{Sketch stroke saliency}
\label{subsec:SketchStrokeSaliency}

We use the agent trained with the proposed  ranked reward  and exploit its output distribution $\phi_{t}$ to compute a saliency value $\mathbb{S}\in [ 0,1 ] $ for each stroke in a sketch as:
\begin{equation}
	\mathbb{S}_l= \frac{ \sum_{t=l_{min}}^{l_{max}} \phi_{t}(a_t = 1)}{l_{max} - l_{min}}
	\label{Eq:Saliency}
\end{equation}
\noindent where $l \in \{1, 2, \cdots L\}$ is the stroke index, $L$ is the total number of strokes in a sketch, $l_{min}$ is the time step $t$ corresponding to the first stroke-segment in the stroke with index $l$ and $l_{max}$ corresponding to the last one. Thus strokes which the agent learns are important to keep for obtaining high recognition (or ranking) accuracy are more salient.

\subsection{Category-level sketch synthesis}
\label{subsec:SketchSynthesis}

Combining our abstraction model with the VAE RNN category-level sketch synthesis model  in \cite{ha2017neural}, we obtain a sketch synthesis model with controllable abstraction.
Specifically, once the synthesizer is trained to generate sketches for a given category,  we  use it to generate a sketch of that category. This is then fed to our abstraction model, which can generate different versions of the input sketch at the desired abstraction level as explained in Sec.~\ref{subsubsec:VariedLevelOfAbstraction}.
% to obtain multiple abstracted versions of the input sketch using the method described in Subsection \ref{subsubsec:VariedLevelOfAbstraction}. \todo{not sure about the fine-tuning part. Actually there is no fine-tuning here.} \Unote{Actually here we didn't do fine-tuning.} %we fine-tune our abstraction model on synthesized sketches by easily integrating the synthesis model in our framework, thanks to its modular nature. At the start of each trajectory iteration, a whole sketch is synthesized which is then fed into the abstraction framework as if it is loaded from a real dataset.

%One of the limitation of this synthesis model is the lack of ability to generate good quality sketches when its trained on multi-class sketch data. We try to overcome this problem by training a separate model for each sketch category. During fine-tuning of the abstraction model, we load all the trained synthesis models and then select randomly one to generate the sketch.

%Once the abstraction model is fine-tuned, sketches can be synthesized at various levels using the method described in Subsection \ref{subsubsec:VariedLevelOfAbstraction}. 

\subsection{Photo to sketch synthesis}
\label{subsec:PhotoToSketchSynthesis}

\begin{figure}[ht]
	\centering
	\includegraphics[width=\linewidth]{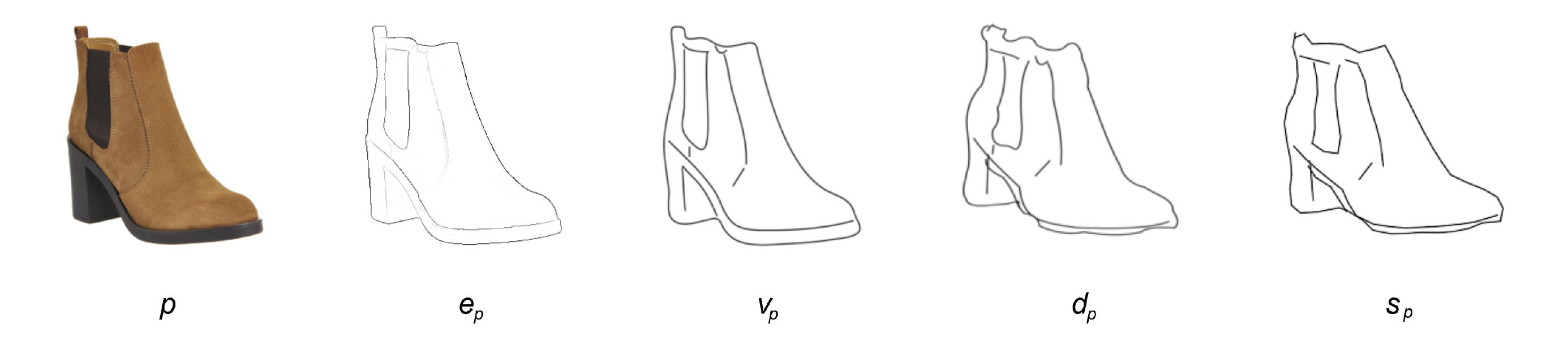}
	\caption{Pre-processing before photo-to-sketch synthesis.}
	\label{fig:EdgeSimplification}
\end{figure}

Based on our abstraction model, we propose a novel photo-to-sketch synthesis model that is completely different from prior cross-domain image synthesis methods \cite{sangkloy2016scribbler, isola2016image}  based on encoder-decoder training. Our approach consists of the following steps (Fig.~\ref{fig:EdgeSimplification}). (1) Given a photo $p$, its edge-map $e_p$ is extracted using an existing edge detection method \cite{ZitnickECCV14edgeBoxes}. (2) We do not use a threshold  to remove the noisy edges as in \cite{ZitnickECCV14edgeBoxes}. Instead, we keep the noisy edge detector output as it is and use a line tracing algorithm {\cite{imagemagick2017llc}} to convert the raster image to a vector format, giving vectorized edge-maps $v_p$. (3) Since contours in human sketch are much less smooth than those in a photo edge-map, we apply non-linear transformations/distortions to $v_p$ both at the stroke and the whole-sketch (global) level. At global-level, these transformations include rotation, translation, rescaling, and skew both along x-axis and y-axis. At stroke-level they include translation and jittering of stroke curvature. After these distortions, we obtain $d_p$, which has rougher contours as in a human free-hand sketch (see Fig.~\ref{fig:EdgeSimplification}). (4) The distorted edge-maps are then simplified to obtain $s_p$ to make them more compatible with the type of free-hand sketch data on which our abstraction model is trained. This consists of fixed-length re-sampling of the vectorized representation to reduce the number of data-segments. (5) After all these preprocessing steps,  $s_p$ is used as input to our abstraction model to generate abstract sketches corresponding to the input photo $p$. Before that, the abstraction model is  fine-tuned on pre-processed edge-maps $s_p$.%The described pre-processing steps can be visualized in Figure \ref{fig:EdgeSimplification}. 
%The code for our photo-to-sketch synthesis model together with code for other models introduced in this paper will be made available in the first author's website to ensure reproducibility. 

%Similarly to what we did in Subsection \ref{subsec:SketchSynthesis}, before actually using our model to generate abstraction representations of photos, we fine-tune it on simplified edge-maps $s_p$.

\subsection{Fine-grained SBIR}
\label{subsec:FGSBIR}

Armed with the proposed sketch abstraction model and the photo-to-sketch synthesis model presented in Sec.~\ref{subsec:PhotoToSketchSynthesis}, we can now train a FG-SBIR given photos only.

\begin{figure}[t]
	\centering
	\includegraphics[width=0.9\linewidth]{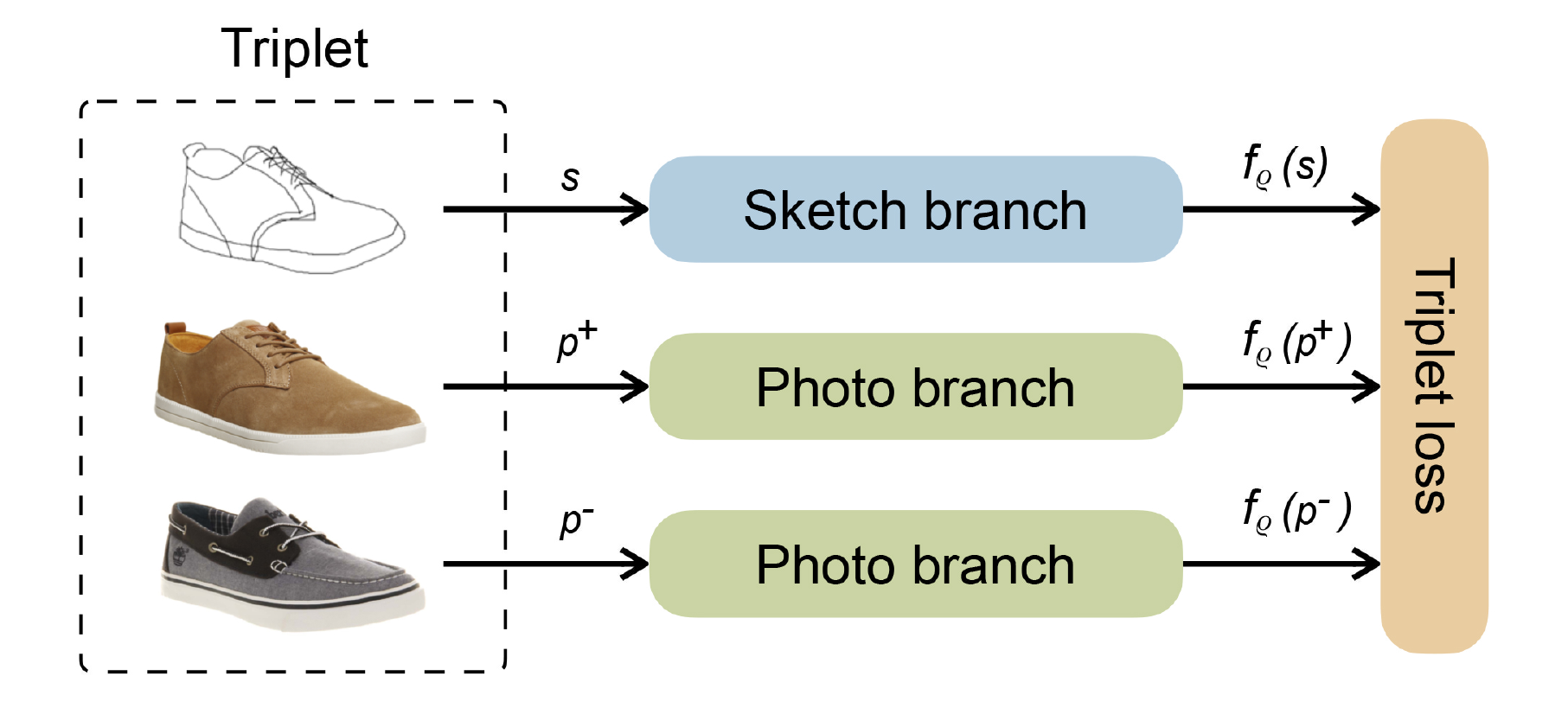}
	\caption{The FG-SBIR model \cite{yu2016sketch}.}
    \vspace{-0.6cm}
	\label{fig:FG-SBIR}
\end{figure}

Given a set of training object photo images, we take each photo $p$ and generate its simplified edge-map $s_p$. This is then fed into the abstraction model to get three levels of abstraction $a^{1}_p$, $a^2_p$ and $a^3_p$, by setting  $\delta$ to   $-0.1$, $0.0$ and $+0.1$ respectively (see Eq.~\ref{Eq:control abstract}). This procedure provides three sketches for each simplified edge-map of a training photo, which can be treated as photo-sketch pairs for training a FG-SBIR model. Concretely, we employ the triplet ranking model  \cite{yu2016sketch}  illustrated in Fig.~\ref{fig:FG-SBIR}.
It is a three-branch Siamese CNN. The input to the model is a triplet including a query sketch $s$,  a positive photo $p^+$ and negative photo $p^-$. The network branches aim to learn a joint embedding for comparing photos and sketch such that the distance between $s$ and $p^+$ is smaller than that between $s$ and $p^-$. This leads to a triplet ranking loss:
\begin{equation}
	\begin{split}
		L_\varrho \left( s, p^+, p^- \right) = \max ( 0,\Delta  + D \left( f_\varrho \left( s \right),f_\varrho \left( p^+  \right) \right) \\
		- D\left( f_\varrho \left( s \right),f_\varrho \left( p^-  \right) \right) )
	\end{split}
\end{equation}
\noindent where $\varrho$ denotes the model parameters, $f_\varrho (\cdot)$ denotes the output of the corresponding network branch, $D(\cdot,\cdot)$ denotes Euclidean distance between two input representations and $\Delta$ is the required margin between the positive query and negative query distance. During training we use $s_p$,  $a^1_p$, $a^2_p$ and $a^3_p$ with various distortions (see Sec.~\ref{subsec:FGSBIRExperiments}) in turn as the query sketch $s$. The positive photo $p^+$ is the photo used to synthesize the  sketches, and the negative photo is any other  training photo of a different object.

During testing, we have a gallery of test photos which have no overlap with the training photos (containing completely different object instances), and the query sketch now is a real human free-hand sketch. To deal with the variable  abstraction in human sketches (see Fig.~\ref{fig:SketchAbstractionLevels}), we also apply our sketch abstraction model to the query test sketch  and generate three abstracted sketches as we did in the training stage. The four query sketches are then fed to the trained FG-SBIR model and the final result is obtained by score-level fusion over the four sketches. % We demonstrate the effectiveness of this model, trained without any photo-sketch pair, by testing on human drawn sketches. 

\begin{figure*}[t]
	\centering
	\includegraphics[width=0.92\textwidth]{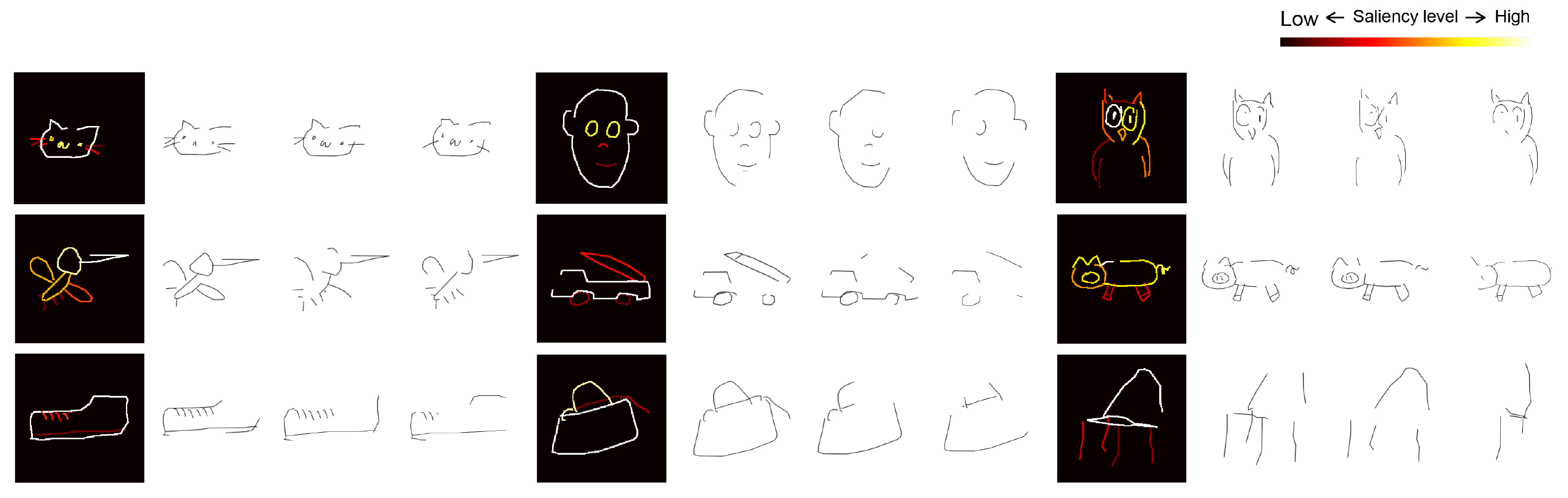}
	%\vspace{-0.1cm}
    \caption{Examples of sketch abstraction and stroke saliency. For each object, the input human sketch annotated with stroke saliency (color coded) computed by model is shown with black background. Three corresponding sketches of  different abstraction level (level 1 to 3, left to right) obtained with our  model are shown with white background. Best viewed in color.}
\vspace{-0.3cm}	\label{fig:AbstractionAndSaliency}
\end{figure*}

\section{Experiments}
\label{sec:Experiments}

\subsection{Sketch abstraction}
\label{subsec:SketchAbstractionExperiments}

\keypoint{Datasets} We use QuickDraw  \cite{ha2017neural}  to train our sketch abstraction model.  It is the largest free-hand sketch dataset to date. % with more than 50 million sketches from 345 categories. 
We select 9 categories (cat, chair, face, fire-truck, mosquito, owl, pig, purse, shoe) with 75000 sketches in each category, using 70000 for training and the rest for testing.

\keypoint{Implementation details} Our code is written in Tensorflow \cite{tensorflow2015-whitepaper}. We implement the B-GRU module of the agent using a single layered B-GRU with 128 hidden cells, which is trained with a learning rate $\eta$ of 0.0001. The RL environment is implemented using  standard step and reset functions. In particular, the step function includes the data updater and reward generator module. The sketch classifier used to generate reward is a three-layer LSTM, each layer containing 256 hidden cells. We train the classifier on the 9 categories using cross-entropy loss and Adam optimizer, obtaining an accuracy of $97.00\%$ on the testing set. The parameters of the ranked reward scheme (see Sec.~\ref{subsubsection:RewardScheme}) are set to: $w_{rf} = 0.5$, $w_c = 0.8$ and $w_v = 0.2$.

\keypoint{Baseline} We compare our abstraction model with random skipping of stroke-segments from each sketch so that the number of retained data-segments is equal in both models.% is equal to the average number of data-segments in the abstracted sketches obtained using our model.

\keypoint{Results} In this experiment, we take the human free-hand sketches in the test set of the 9 selected QuickDraw categories and generate three versions of the original sketches with different abstraction levels. These are obtained by   setting the model parameter $\delta$ to   $-0.1$, $0.0$ and $+0.1$ respectively (Eq.~\ref{Eq:control abstract}).  Some qualitative  results are shown in Fig.~\ref{fig:AbstractionAndSaliency}. It can be seen that the abstracted sketches preserve the most distinctive parts of the sketches. For quantitative evaluation, we feed the three levels of abstracted sketches to the sketch classifier trained using the original sketches in the training set and obtain the recognition accuracy. The results in Table~\ref{tab:Abstraction} show that the original sketches in the test set has 64.79 data segments on average. This is reduced to 51.31, 43.33, and 39.48 using our model with different values of $\delta$.  Even at the abstraction level 3 when around 40\% of the original data segments have been removed, the remaining sketches can still be recognized at a high accuracy of 70.40\%.  In contrast, when similar amount of data segments are randomly removed (Baseline), the accuracy is 6.20\% lower at 64.20\%. This shows that the model has learned which segments can be removed with least impact on recognizability. Table \ref{tab:Abstraction} also compares the proposed ranked reward scheme (Eq.~\ref{Eq:rRank}) with the Basic Reward  (Eq.~\ref{Eq:ra}). It is evident that the ranked reward scheme is more effective. 

\begin{table}[t]
	\centering
	\resizebox{\linewidth}{!}{%
		\begin{tabular}{ll|cc}
			& \multicolumn{1}{c}{} & \multicolumn{1}{c}{\#DataSegments} & Accuracy \\ \hline
			\multicolumn{2}{c|}{Full Sketch} & 64.79 & 97.00\% \\ \hline
			\multicolumn{1}{l|}{\multirow{3}{*}{\begin{tabular}[c]{@{}c@{}}1st Level Abstraction \\ ($\delta=-0.1$)\end{tabular}}}  & Baseline & 51.00 & 85.00\% \\  
			\multicolumn{1}{l|}{} & Basic Reward & 51.12 & 87.60\% \\
			\multicolumn{1}{l|}{} & Ranked Reward & 51.31 & \textbf{88.20}\% \\ \hline
			
			\multicolumn{1}{l|}{\multirow{3}{*}{\begin{tabular}[c]{@{}c@{}}2nd Level Abstraction \\ ($\delta=0.0$)\end{tabular}}} & Baseline & 43.00 & 74.60\% \\ 
			\multicolumn{1}{l|}{} & Basic Reward & 43.09 & 78.80\% \\ 
			\multicolumn{1}{l|}{} & Ranked Reward & 43.33 & \textbf{80.80}\% \\ \hline
			
			\multicolumn{1}{l|}{\multirow{3}{*}{\begin{tabular}[c]{@{}c@{}}3rd Level Abstraction \\ ($\delta=+0.1$)\end{tabular}}} & Baseline & 39.00 & 64.20\% \\
			\multicolumn{1}{l|}{} & Basic Reward & 39.37 & 68.00\% \\ 
			\multicolumn{1}{l|}{} & Ranked Reward & 39.48 & \textbf{70.40}\% \\ \hline
		\end{tabular}%
	}
	%	\vspace*{+0.5mm}
	\caption{Recognizability of abstracted human sketches.}
    \vspace{-0.6cm}	
	\label{tab:Abstraction}
\end{table}

\keypoint{Measuring sketch stroke saliency}
Using Eq.~\ref{Eq:Saliency}, we can compute a saliency value  $\mathbb{S}$ for each stroke in a sketch, indicating how it contributes towards the overall recognizability of the sketch. Some example stroke saliency maps obtained on the test set are shown in Fig.~\ref{fig:AbstractionAndSaliency}. We observe that high saliency strokes correspond to the more distinctive visual characteristics of the object category. For instance, for shoe, the overall contour is more salient than the shoe-laces because many shoes in the dataset do not have shoe-laces. Similarly, for face, the outer contour is the most distinctive part, followed by eyes and then nose and mouth -- again, different people sketch the nose and mouse very differently;  but they are more consistent in drawing the outer contour and eyes. These results also shed some light into how deep sketch recognition models make their decisions, providing an alternative to gradient-based classifier-explanation approaches such as \cite{selvaraju2017gradCam}.
%These results shed some light into how human perceive an objects and could be exploited for improving a recognition model by focusing on the more salient parts. 

\subsection{Sketch synthesis}
\label{subsec:SketchSynthesisExperiments}

\begin{figure}[t]
	\centering
	\includegraphics[width=0.99\linewidth]{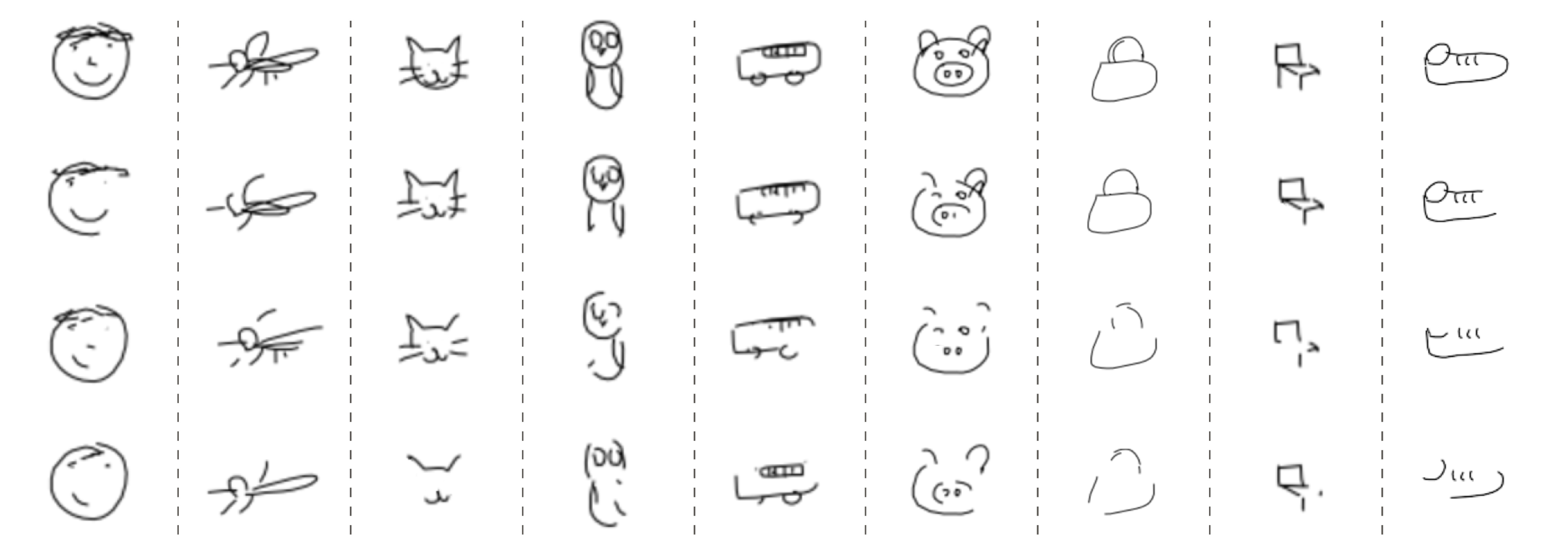}
	\vspace{-0.1cm}
	\caption{Examples of synthesized sketches at different abstraction levels. Top to bottom: increasing abstraction levels.}
\vspace{-0.6cm}	\label{fig:SketchSynthesisEx}
\end{figure}

We train a sketch synthesis model as in \cite{ha2017neural} for each of the 9 categories, and combine it with our  abstraction model (Sec.~\ref{subsec:SketchAbstractionExperiments}) to generate abstract versions of the synthesized sketches. Again, we compare our abstraction results with the same random removal baseline. 
From the quantitative results in Table \ref{tab:AbstractSketchSynthesis}, we can draw the same set of conclusions: the synthesized sketches are highly recognizable even at the most abstract level, and more so than the sketches generated with random segment removal.   Fig.~\ref{fig:SketchSynthesisEx} shows some examples of synthesized sketches at different abstraction levels. 

\begin{table}[t]
	\centering
	\resizebox{\linewidth}{!}{%
		\begin{tabular}{ll|cc}
			& \multicolumn{1}{c}{} & \multicolumn{1}{c}{\#DataSegments} & Accuracy \\ \hline
			\multicolumn{2}{c|}{Full Sketch} & 69.61 & 99.6\% \\ \hline
			\multicolumn{1}{l|}{\multirow{3}{*}{\begin{tabular}[c]{@{}c@{}}1st Level Abstraction \\ ($\delta=-0.1$)\end{tabular}}} & Baseline & 50.00 &  89.96\% \\  
			\multicolumn{1}{l|}{} & Basic Reward & 50.43 & 92.60\% \\
			\multicolumn{1}{l|}{} & Ranked Reward &  50.08 & \textbf{94.20}\% \\ \hline
			\multicolumn{1}{l|}{\multirow{3}{*}{\begin{tabular}[c]{@{}c@{}}2nd Level Abstraction \\ ($\delta=0.0$)\end{tabular}}} & Baseline & 44.00 & 80.20\% \\ 
			\multicolumn{1}{l|}{} & Basic Reward & 44.13 & 88.40\% \\ 
			\multicolumn{1}{l|}{} & Ranked Reward & 44.32 & \textbf{90.80}\% \\ \hline
			\multicolumn{1}{l|}{\multirow{3}{*}{\begin{tabular}[c]{@{}c@{}}3rd Level Abstraction \\ ($\delta=+0.1$)\end{tabular}}} & Baseline & 37.00 & 69.20\% \\
			\multicolumn{1}{l|}{} & Basic Reward & 37.15 & 73.20\% \\ 
			\multicolumn{1}{l|}{} & Ranked Reward & 37.56 & \textbf{79.40}\% \\ \hline
		\end{tabular}%
	}
	%	\vspace*{+0.5mm}
	\caption{Recognizability of category-level synthesized sketches.}
	\label{tab:AbstractSketchSynthesis}
\end{table}

\subsection{Photo to sketch synthesis}
\label{subsec:PhotoToSketchSynthesisExperiments}

\keypoint{Dataset} We use the QMUL Shoe-V2 dataset \cite{qmulshoechair2017}.  It is the largest single-category FG-SBIR dataset  with 1800 training and 200 testing photo-sketch pairs. %Only the training photos are used in training our model.

\keypoint{Implementation details} As described in Sec.~\ref{subsec:PhotoToSketchSynthesis}, we fine-tune our abstraction model, previously trained on the 9 classes of QuickDraw dataset, on the simplified edge-maps $s_p$ of the training photos from Shoe-V2.

\keypoint{Baseline} We compare our model with our implementation of the cross-domain deep encoder-decoder based  synthesis model in  \cite{sangkloy2016scribbler}. Note that although it is designed for synthesis across any direction between photo and sketch, only sketch-to-photo synthesis results are shown in \cite{sangkloy2016scribbler}. 

\begin{figure}[t]
	\centering
	\includegraphics[width=0.88\linewidth]{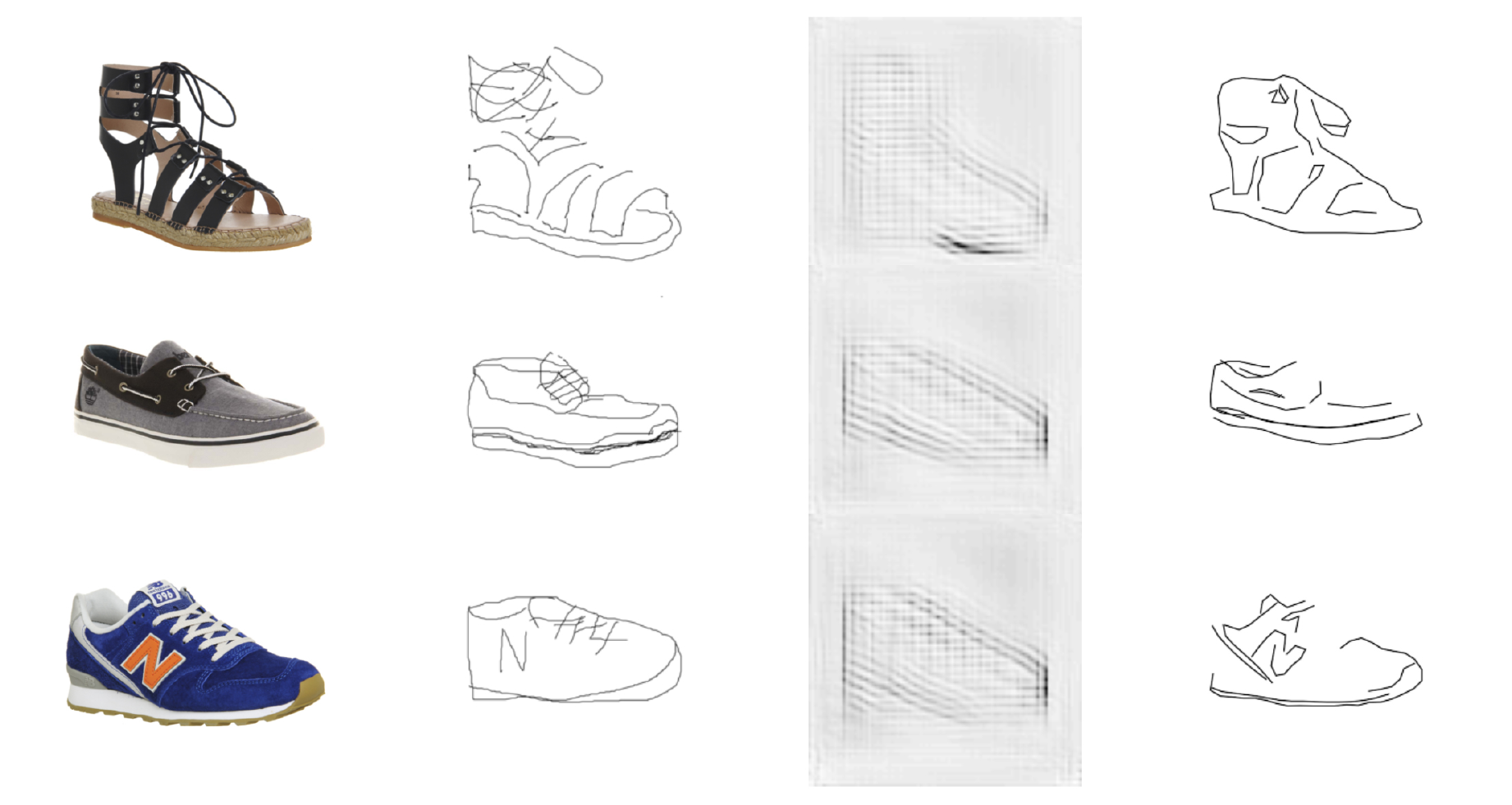}
	\vspace{-0.1cm}
	\caption{Examples of synthesized sketches using \cite{sangkloy2016scribbler} (third col) and ours (fourth) vs human sketch (second).}
\vspace{-0.6cm}	\label{fig:Photo2SketchSynthesis}
\end{figure}

\keypoint{Results} We show some examples of the synthesized sketches using our model and  \cite{sangkloy2016scribbler} in Fig.~\ref{fig:Photo2SketchSynthesis}. We  observe that our model produces much more visually appealing  sketches than the ones obtained using \cite{sangkloy2016scribbler}, which is very blurry and seems to suffer from mode collapse. This is not surprising: the dramatic domain gaps and the mis-alignment between photo and sketch makes a deep encoder-decoder model such as \cite{sangkloy2016scribbler} unsuitable. Furthermore, treating a sketch as a 2D matrix of pixels is also inferior to treating it as a vectorized coordinate list as in our model.

\subsection{Fine-grained SBIR}
\label{subsec:FGSBIRExperiments}

\keypoint{Dataset} Apart from Shoe-V2, we also use  QMUL Chair-V2, with 200 training  and 158 testing photo-sketch pairs.

\keypoint{Implementation details} As described in Sec.~\ref{subsec:PhotoToSketchSynthesisExperiments}, we generate 5 distortion representations $d_p^{m}$, $m \in \{1,2,3,4,5\}$, for each input vectorized edge-map $v_p$. We then use all $a_p^{m, n}$ representations and simplified edge-maps $s_p^{m}$ to train the state of the art FG-SBIR model  \cite{yu2016sketch}. %During the testing phase we perform abstraction on sketches as well, using the same $\delta$ values we three abstract sketch representations $a_s^m$. We perform the score fusion of $a_s^m$ scores to obtain the final score for the input sketch $s$.

\keypoint{Baseline} Apart from comparing with the same model \cite{yu2016sketch} trained with the annotated photo-to-sketch pairs (`Upper Bound'), we compare with two  baselines using the same FG-SBIR model but trained with different synthesized sketches. Baseline1 is trained with synthesized sketches using the model in \cite{sangkloy2016scribbler}. Baseline2 uses the simplified edge-maps $s_p^{m}$ directly as replacement for human sketches. 

\begin{figure}
	\centering
	\includegraphics[width=0.9\linewidth]{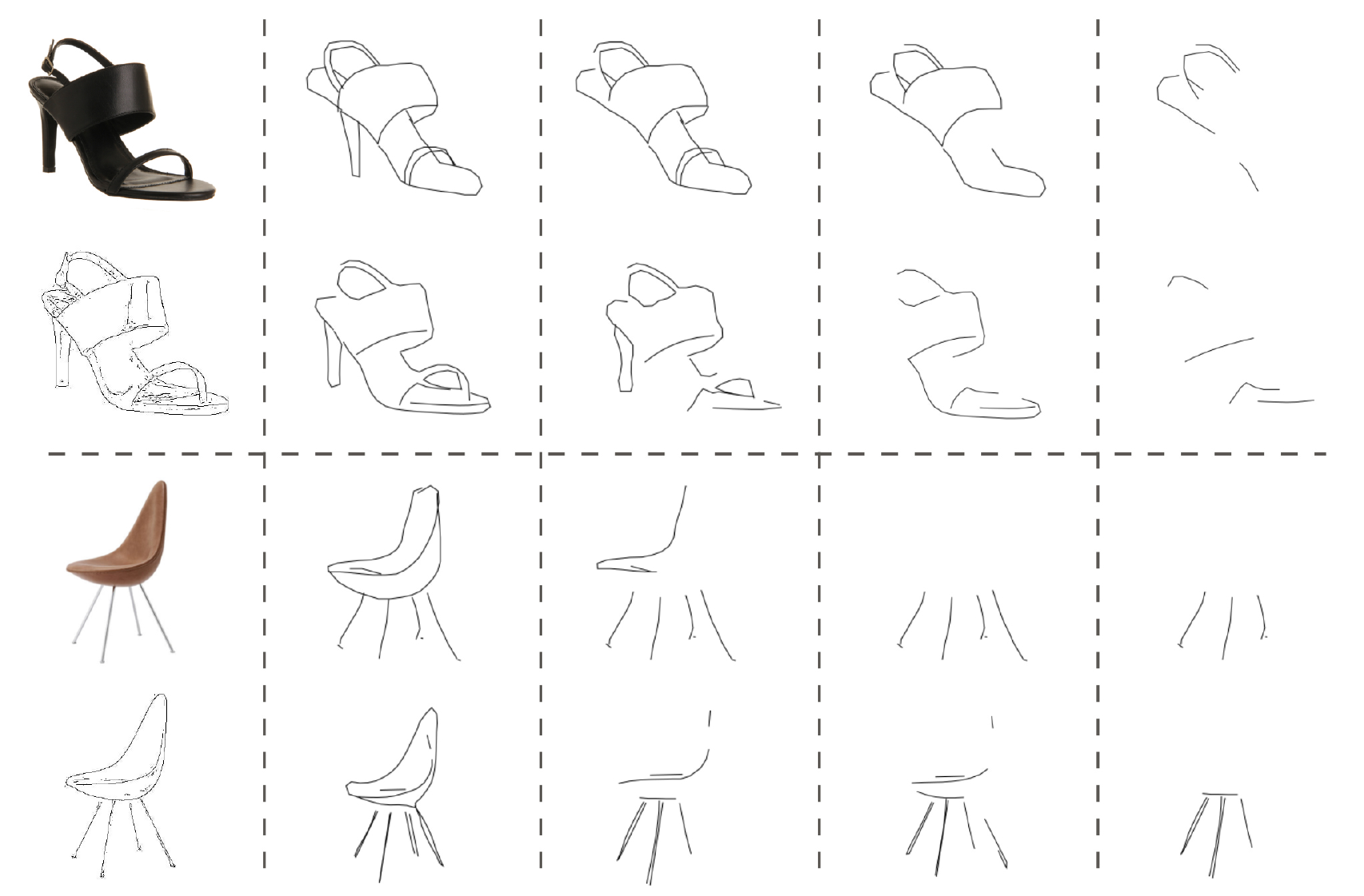}
	\vspace{-0.1cm}	
	\caption{Human and synthesized sketches at different abstraction level used in the FG-SBIR experiments. For each object: First row: photo, sketch and the abstracted sketches. Second row: edge-map and  synthesized sketches.}
    \vspace{-0.6cm}	
	\label{fig:SBIRexamples}
\end{figure}

\keypoint{Results} Table \ref{tab:SBIRtable} shows that the model trained with synthesized sketches from our photo-to-sketch synthesizer is quite competitive, e.g., on chair, it is only 7.12\% lower on Top 1 accuracy.  It decisively beats the model trained with sketches synthesized using \cite{sangkloy2016scribbler}. The gap over Baseline2 indicates that the abstraction process indeed makes the generated sketches more like the human sketches. Some qualitative results are shown in Fig.~\ref{fig:SBIRexamples}. Note the visual similarity between synthesized sketches at different abstraction levels and the corresponding abstracted human sketches. They are clearly more similar at the more abstract levels, explaining why it is important to include sketches at different abstraction levels during both training and testing. 

\begin{table}[ht]
	\centering
	\resizebox{\linewidth}{!}{%
		\begin{tabular}{l|cc|cc}
			%\cline{5-8}
			\multicolumn{1}{c}{}& \multicolumn{2}{c}{Shoe-V2} & \multicolumn{2}{c}{Chair-V2} \\ \hline
			Method & Top1 & Top10 & Top1 & Top10 \\ \hline \hline 
			Baseline1 \cite{sangkloy2016scribbler} & 8.86\% & 32.28\% & 31.27\% & 78.02\% \\ 
			Baseline2 & 16.67\% & 50.90\% & 34.67\% & 73.99\% \\ 
			Ours & \textbf{21.17\%} & \textbf{55.86\%} & \textbf{41.80\%} & \textbf{84.21\%} \\ \hline \hline
			Upper Bound& 34.38\% & 79.43\% & 48.92\% & 90.71\% \\  \hline
		\end{tabular}%
	}
	%	\vspace*{+0.5mm}
	\caption{FG-SBIR results. Top 1 and 10 matching accuracy. }
	\label{tab:SBIRtable}
\end{table}

\vspace{-5mm}
\subsection{Human Study}
\label{subsec:HumanStudy}
\vspace{-1mm} In this study, 10 users were shown 100 pairs of abstracted sketches from the same 9 classes used in Sec.~\ref{subsec:SketchAbstractionExperiments}. Each pair consists of a sketch obtained using our framework and another sketch obtained by randomly removing stroke-segments. Each pair is shown side by side and the relative position of the two sketches is random to prevent any bias. The users were asked to choose the more aesthetically appealing sketch among each pair. Results in percentage (Mean: 64.3 $\pm$ 4.59, Min: 58, Max: 70) suggest that the abstracted sketches produced by our model are more visually appealing to humans when compared with sketches with randomly removed stroke-segments.

% \begin{table}[h]
% 	\centering
% 	\begin{tabular}{c|c|c}
% 		\hline
% 		Mean $\pm$ Std & Min & Max  \\
% 		\hline
% 		64.3 $\pm$ 4.59 & 58 & 70\\
% 		\hline
% 	\end{tabular}%
% 	\vspace{0.05cm}
% 	\caption{User study of abstracted sketch aesthetics. Percentage of times our abstracted sketches were preferred over the random stroke removal baseline.}
% 	\label{tab:HumanStudy}
% \end{table}

% The results, shown in Table~\ref{tab:HumanStudy}, 

\vspace{-2mm}
\section{Conclusion}
% Future work...extend this to edgemaps in the wild.
\vspace{-1mm} We have for the first time proposed a stroke-level sketch abstraction model. Given a sketch, our model learns to predict which strokes can be safely removed without affecting overall recognizability. We proposed a reinforcement learning framework with a novel rank-based reward to enforce stroke saliency. We showed the model can be used to address a number of existing sketch analysis tasks. In particular, we demonstrated that a FG-SBIR model can now be trained with photos only. In future work we plan to make this model more practical by extending it to work with edge-maps in the wild. We also intend to develop an end-to-end trained abstraction model which could directly sample a variable abstraction-level sketch.

{\small
	\bibliographystyle{ieee}
	\bibliography{egbib}
}
	
\end{document}